\documentclass[letterpaper, 10 pt, conference]{ieeeconf}  

\IEEEoverridecommandlockouts                              

\usepackage{algorithm,algpseudocode}
\usepackage{wrapfig}
\usepackage{lipsum}
\usepackage{subcaption}
\usepackage[utf8]{inputenc} 
\usepackage[T1]{fontenc}    
\usepackage{hyperref}       
\usepackage{url}            
\usepackage{booktabs}       
\usepackage{amsfonts}       
\usepackage{nicefrac}       
\usepackage{microtype}      
\usepackage{lipsum}
\usepackage{algorithm,algpseudocode}
\usepackage{amsmath}
\usepackage{mathtools}
\usepackage{subcaption}
\usepackage{algpseudocode}
\usepackage{setspace}
\usepackage{subcaption}
\usepackage{makecell}
\usepackage{mleftright}
\usepackage{xcolor}
\usepackage[noadjust]{cite}

\DeclareMathSymbol{@}{\mathord}{letters}{"3B}

\newcommand{\mypara}[1]{\vspace{2pt}\noindent\textbf{#1}}

\newcommand{\vect}[1]{\boldsymbol{#1}}
\newcommand{\argmax}{\operatornamewithlimits{argmax}}
\newcommand{\comment}[1]{}
\DeclareMathOperator{\E}{\mathbb{E}}

\DeclareMathSymbol{@}{\mathord}{letters}{"3B}
\newcommand{\Sha}[1]{\textcolor{black}{#1}}
\newcommand{\sh}[1]{\textcolor{black}{#1}}

\title{\LARGE \bf
Neural fidelity warping for efficient robot morphology design
}

\author{Sha Hu$^{1}$,  Zeshi Yang$^{1}$,  and Greg Mori$^{1}$
\thanks{$^{1}$School of Computing Science, Simon Fraser University, BC, CA.
{\tt\small hushah@sfu.ca, zeshiy@sfu.ca, mori@cs.sfu.ca}}%
}

\begin{document}

\maketitle
\thispagestyle{empty}
\pagestyle{empty}

\begin{abstract}

We consider the problem of optimizing a robot morphology to achieve the best performance for a target task, under computational resource limitations. The evaluation process for each morphological design involves learning a controller for the design, which can consume substantial time and computational resources. To address the challenge of expensive robot morphology evaluation,  
we present a continuous multi-fidelity Bayesian Optimization framework that efficiently utilizes computational resources via low-fidelity evaluations.  We identify the problem of non-stationarity over fidelity space. Our proposed fidelity warping mechanism can learn representations of learning epochs and tasks to model non-stationary covariances between continuous fidelity evaluations which prove challenging for off-the-shelf stationary kernels. Various experiments demonstrate that our method can utilize the low-fidelity evaluations to efficiently search for the optimal robot morphology, outperforming state-of-the-art methods.

\end{abstract}

\section{INTRODUCTION}
A longstanding goal in robotics~\cite{brooks2000robot} is the automatic construction of robots that are functional in the real world. To realize this grand goal, full autonomy should be achieved not only at the level of robot control, but also at the level of robot design.  However, compared with recent successes in automatic robot control, the progress towards automatic robot design is limited.

The robot design problem is challenging. It requires efficient search over a large robot design space. Furthermore, precise robot design evaluation requires obtaining a controller and then executing actions to perform a task. Learning a deep reinforcement learning based controller is computation-intensive and time-consuming; thus it is expensive to conduct iterations of precise evaluation to search for an optimal design.  Designers often use cheap simulators to get a sense of how a robot model would perform in the real world. Inspired by the fact that cheap yet inaccurate evaluations can unravel a  "black box" problem, we propose leveraging low fidelity levels of evaluation to optimize a robot morphology. Instead of evaluating the performance on a difficult task (e.g., long time horizon, complex scenarios), we can use easier tasks (shorter time horizon, simple scenarios) to test a candidate robot design. Similarly, we can reduce the time used to train the controller. Performance on easier tasks and performance on a target difficult task are related because they share a similar reward function and state transition probability.  Similarly, performance with a policy learned from fewer epochs of learning is related to that from many epochs because they share a controller of the same architecture. Therefore, the knowledge we obtain by evaluating one design on easier tasks or with fewer epochs can help identify optimal morphology on the target fidelity.  These multi-fidelity evaluations enable us to explore the robot morphology space efficiently.

\begin{figure}
\centering
\includegraphics[width=1\linewidth]{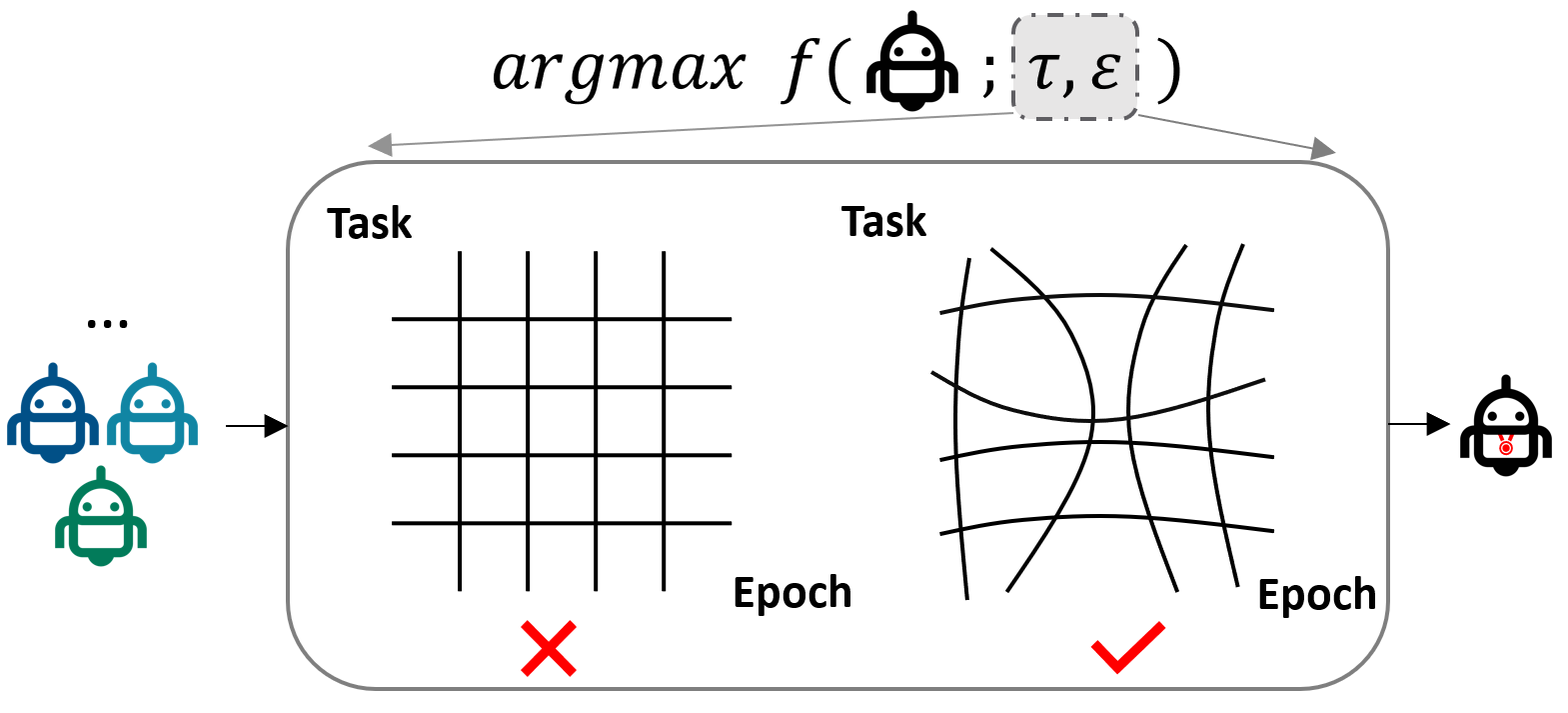}
\caption{Our goal is to efficiently search an optimal robot morphology. By utilizing easier tasks and fewer epochs to evaluate candidate morphologies, we can efficiently explore the morphology space. However, the correlations between the evaluations by different fidelities cannot be captured by the distance measured in the raw fidelity space. We present an approach to model the complex relationships across fidelities.}
\label{fig: pullfigure}
\end{figure}

We ground this multi-fidelity idea in a Bayesian Optimization (BO) framework. BO is not immediately a practical framework for the automatic robot design task due to  extremely costly evaluation. Multi-fidelity BO (MFBO) has emerged as a cost-efficient method for solving expensive evaluations and has achieved impressive success in many applications such as hyperparameter tuning for machine learning algorithms~\cite{swersky2013multi, kandasamy2017multi, klein2017fast, song2019general, takeno2019multi}. A key component that makes MFBO effective is to accurately estimate the relations between evaluations from different fidelities so that information from low-fidelity evaluations can be aggregated to estimate the target-fidelity objective. The correlations between evaluations at different fidelities are complex. However, existing MFBO approaches assume the correlation is invariant to translations in fidelity space using a stationary kernel~\cite{kandasamy2017multi} or impose a monotonic correlation structure by simple basis function~\cite{klein2017fast}. Hence, they can be less accurate in estimating the objective function, leading to suboptimal return and higher search cost. 

In this work, we propose a simple yet effective method to address the non-stationarity across the fidelity space.  Specifically, we use a neural network based warping function to project the raw representations of the task and number of epochs to new representations so that the new representations can be modeled with a standard stationary kernel. Our method is flexible to capture the complex relationships between the fidelities, and all the posterior computation remains tractable. Our algorithm efficiently achieves better robot design optimization results.


\section{Related Work}
\mypara{Automatic Robot Morphology Design.}
The morphology of a robot imposes a prior on the capacity and learning behavior of the robot. The pivotal role of morphology has been well acknowledged~\cite{pfeifer2006body}, and studied under various methodologies. Pioneering works include evolutionary computation~\cite{sims1994evolving, sims1994evolving1, funes1998evolutionary, lipson2000automatic, bongard2003evolving, paul2001road}. \cite{sims1994evolving, sims1994evolving1} use an Evolutionary Strategy (ES) to train a neural network based controller, and mutate the robot structure graph after each generation. \cite{funes1998evolutionary} build physical versions of the generated robots from ES by Lego blocks, and~\cite{lipson2000automatic} use rapid manufacturing technology to fabricate the evolved robots.
~\cite{geijtenbeek2013flexible} employ a Covariance Matrix Adaptation Evolution Strategy to optimize over both the motion control and muscle attachment points. However, evolutionary approaches have limited guarantees on optimality. Instead, our approach is based on a global optimization algorithm that can efficiently explore design spaces.

Several other approaches have been explored in the direction of evolutionary robotics. \cite{spielberg2017functional} propose using a sparse sequential quadratic programming solver to optimize the robot's physical and trajectory parameters jointly. However, it requires a robot designer to provide initial guesses over those parameters to achieve a feasible solution. \cite{ha2017joint, ha2018computationalb} apply the implicit function theorem to derive the relationships among the design and trajectory parameters. \cite{baykal2017asymptotically} integrate motion planning into Adaptive Simulated Annealing search over the robot design space and provide a theoretical analysis of the asymptotic optimality of their method. Different from all the above methods that adopt trajectory optimization to control the robots, we adopt a parametric control policy, with the aim of studying robot morphology design with powerful deep learning based robot control.

Recent learning based approaches of co-optimizing a robot's morphology and its control include~\cite{luck2020data, wang2019neural, schaff2019jointly, ha2019reinforcement, spielberg2019learning}. \cite{luck2020data} propose using a learned morphology conditioned action value function as the surrogate objective to estimate the performance of the candidate designs to avoid executing a large number of episodes. The estimated performances are then used to guide the design optimization by a Particle Swarm Optimizer. \cite{wang2019neural} utilize a graph structured policy to enable fast policy learning during morphology search. \cite{schaff2019jointly} treat the design parameters as a context variable and feed it to the policy network together with the robot state. The parametric design distribution is then optimized by the policy gradient, similar to the policy optimization. \cite{ha2019reinforcement} introduces the population-based policy gradient method to update the design distribution and policy distribution simultaneously. \cite{spielberg2019learning} learn a low-dimensional latent  representation for robots' states via a variational autoencoder to simultaneously optimize the controller and material parameters. However, none of these methods addresses the issue of expensive evaluation per design.


There also has been a large body of research on modular robots \cite{yim2007modular, zykov2007evolved, romanishin2013m, icer2017evolutionary, mehta2014cogeneration, pathak2019learning, wright2007design, ha2018computationala, yim2000polybot, whitman2020modular, campos2019task, desai2017computational, carlone2019robot} where a set of predefined reusable modules is designed to compose versatile robotic systems to solve a wide variety of tasks on the fly. \cite{pathak2019learning, whitman2020modular} formulate the design search as a Markov Decision Process. \cite{pathak2019learning} learn to perform composing actions such as "link" and "unlink" through deep reinforcement learning. \cite{whitman2020modular} train an action value network to predict the expected return of different assemblies to plan which module to add next. \cite{desai2017computational} develop a platform that enables a human-in-the-loop iterative design process for customized assembly. 
\cite{ha2018computationala} formulate the design process as a path planning problem and propose to use A* with a heuristic function to guide design search.   However, those approaches for modular robot design solve discrete optimizations over the combinatorial set of possible arrangements of given finite modules. In contrast, the method we propose searches over continuous design space and can produce robots of novel components entirely from scratch.

\mypara{Multi-fidelity Bayesian Optimization.}
Bayesian Optimization (BO)~\cite{jones1998efficient, shahriari2015taking} is a well-studied global optimization method for solving expensive-to-evaluate black-box optimization problems with relatively few function evaluations. BO has been used for robot pushing~\cite{wang2017focused}, bipedal locomotion~\cite{calandra2014experimental}, and robot path planing~\cite{marchant2014bayesian}.
\cite{liao2019data} is a recent work adopting BO to address the robot morphology design problem. It employs Batch BO to enable parallel evaluations for a batch of morphologies in the application settings where parallel manufacture is available.  Multi-fidelity BO (MFBO)~\cite{swersky2013multi, kandasamy2017multi, klein2017fast, song2019general, takeno2019multi} has been proposed for solving a computationally expensive primary problem by actively querying a cheap yet relevant task. The desiderata for robot morphology design include sufficient explorations over the design space without paying huge expense. We adopt the basic multi-fidelity framework of~\cite{klein2017fast} as a starting point for our method, and develop it further to accommodate the robot morphology design setting, and instill fidelity space warping capability to model the complex relations between evaluations from different fidelities.  

\section{Preliminaries}
\subsection{Problem Formulation}

In this work, we address the robot morphology design problem \Sha{under computational resource limitations}, where the goal is to search over the robot morphology space $\mathcal{X} \subseteq  \mathbb{R}^{n}$ to find the optimal morphology $\vect{x}^{\ast} \in \mathcal{X}$ with $\varepsilon$ epochs to learn a policy for performing task $\tau$. \Sha{The cost to evaluate a robot morphology is a function of $\tau$ and $\varepsilon$, i.e., $c(\tau, \varepsilon)$.} We formally define the robot morphology evaluation function $f: \mathcal{X} \rightarrow \mathbb{R} $ as follows: 
\begin{equation} 
\label{eq:objective function}
f(\vect{x}; \tau, \varepsilon ) = \sum_{t} \E_{(\vect{s}_t,\vect{a}_t) \sim \rho _{ \pi_{\vect{\theta}}^{\tau, \varepsilon}, \vect{x}, \tau}} \left [   r(\vect{s}_t, \vect{a}_t; \vect{x}) \right ],
\end{equation} 
where  $\pi_{\vect{\theta}}^{\tau, \varepsilon}$ is a deep neural network parameterized policy and the parameters are denoted as $\theta$. The superscript $(\tau, \varepsilon)$ means  $\pi_{\vect{\theta}}^{\tau, \varepsilon}$ is obtained after training epochs of $\varepsilon$ under the task $\tau$. $\rho _{\pi_{\vect{\theta}}^{\tau, \varepsilon},\vect{x}, \tau}$ is the state-action marginals of the trajectory distribution induced by the policy $\pi_{\vect{\theta}}^{\tau, \varepsilon}$ conducted by the morphology $\vect{x}$ under the task $\tau$. The task environment emits a reward $r(\cdot; \vect{x})$ on each state-action transition pair $\left ( \vect{s}_t, \vect{a}_t \right )$, and the performance of one morphology $\vect{x}$ is defined as the expected sum of rewards. Therefore, the goal of robot morphology design is expressed as:
\begin{equation} 
\label{eq:RMD}
 \vect{x^{\ast}} = \operatorname*{argmax}_{\vect{x} \in \mathcal{X} } f(\vect{x};\tau, \varepsilon).
\end{equation}
There are two important features of the robot morphology evaluation function $f$.
Firstly, $f$ is considered as a "black-box" function, we do not have access to its gradients due to unknown stochasticity (e.g., unknown state transition probability). We can only access $f(\cdot)$ by observing the outcome of $f(\vect{x})$ by evaluating $\vect{x}$. Secondly, the evaluation process of $\vect{x}$ is expensive. The quality of one morphology $\vect{x}$ is measured by how $\vect{x}$ behaves in the target task, thus to compute the expected return a policy that controls the robot should be learned, which is a compute-intensive process. Therefore we formulate the morphology design problem as expensive-to-evaluate black-box optimization.


\subsection{Bayesian Optimization with Gaussian processes}
\label{sec: bogp}
Bayesian Optimization (BO) is a popular approach for optimizing black-box functions that are expensive to evaluate. Among the components of BO, the \textit{surrogate model} is a critical one to fit observed information and predict outcomes of unseen data. Guassian Processes (GPs)~\cite{rasmussen2003gaussian} are typically used as the \textit{surrogate model} because GPs are a powerful and tractable prior distribution over the objective function. 
By utilizing the posterior predicted distribution over evaluations of unseen data, BO is effective at exploiting previous evaluations and exploring promising regions to identify optima.

A GP model defines a distribution over functions of the form $f: \mathcal{X} \rightarrow \mathbb{R} $, and it is characterized by a mean function $m(\vect{x}): \mathcal{X} \rightarrow \mathbb{R}$ and a covariance function $k(\vect{x}, \vect{x}'): \mathcal{X}^{2} \rightarrow \mathbb{R}$. $m(\vect{x})$ specifies the expected function value at a given $\vect{x}$, and $k(\vect{x}, \vect{x}')$ encodes how the function values of $\vect{x}$ and $\vect{x}{'}$ are correlated. A common choice of the kernel is the anisotropic radial basis function (ARBF) kernel:
\begin{equation}
    k(\vect{x}, \vect{x}{'}) = \theta_{1} \mathrm{exp}(|\vect{x}-\vect{x{'}}|^{\top} M |\vect{x}-\vect{x{'}}|),
\label{equa: RBF-ARD}
\end{equation}
where $\theta_{1}$ represents the magnitude of variation of function values. $M$ is a
diagonal matrix of hyperparameters termed length scale and captures the sensitivity of the function value with respect to changes in~$\vect{x}$. We denote the parameters of the kernel as $\vect{\theta}_k=[\theta_{1}, M]$.

Given a history of $T$ evaluations, denoted as $\mathcal{D}_{T}=\left \{ \left (\vect{x}_{t}, y_{t} \right ) \right \}_{t=0}^{T}$ where $y_{t} = f(\vect{x}_{t}) + \upsilon_{t} 
\in\mathbb{R}$ and $ \upsilon_{t} \sim \mathcal{N}(0, \sigma_{noise}^{2})$, the GP model has an analytical posterior prediction for an unseen data point $\vect{x}_{\star}$ conditioned on previous evaluations and kernel parameters: $y_{\star}|\mathcal{D}_{T}, \vect{\theta}_{k} \sim \mathcal{N}(u_\star, \sigma_\star^{2})$, and the posterior predictive mean and posterior predictive variance are respectively:
\begin{equation}
\begin{gathered}
u_{\star}(\vect{x}_{\star})|\mathcal{D}_{T}, \vect{\theta}_k = \vect{k}^{\top}(\vect{K} + \sigma_{noise}^{2}\vect{I})^{-1}\vect{Y}, \\ \sigma_\star^{2}(\vect{x}_{\star}) |\mathcal{D}_{T}, \vect{\theta}_k =k(\vect{x}_{\star} , \vect{x}_{\star}) -\vect{k}^{\top}(\vect{K} +\sigma_{noise}^{2}\vect{I})^{-1}\vect{k}, 
\end{gathered}
\label{equa: posterior distribution}
\end{equation}
where $\vect{k}\in \mathbb{R}^{T}$ is a vector with $\vect{k}_{t} = k(\vect{x}_{\star}, \vect{x}_{t})$, $\vect{Y} \in \mathbb{R}^{T}$ with $\vect{Y}_{t}= y_{t}$, and $\vect{K}\in \mathbb{R}^{T \times T}$ with $\vect{K}_{i,j} = k(\vect{x}_{i}, \vect{x}_{j})$.

An acquisition function built upon the posterior prediction is used as the selection criterion to determine the $(T +1)^{th}$ robot morphology to evaluate.  We use the information-theoretic acquisition function~\cite{hennig2012entropy,hernandez2014predictive,wang2017max}, due to its global utility measure and relatively fewer hyper-parameters.


However, directly applying BO for robot morphology design is problematic. BO is sample efficient with respect to the number of evaluations needed to achieve an optimal design, but BO still suffers from the expensive cost of robot morphology evaluations. In the following sections, we will present a novel multi-fidelity method which can reduce the required cost of each evaluation.

\section{Neural fidelity Warping for Multi-fidelity Bayesian Optimization}

\begin{figure}
\centering
\includegraphics[width=1\linewidth]{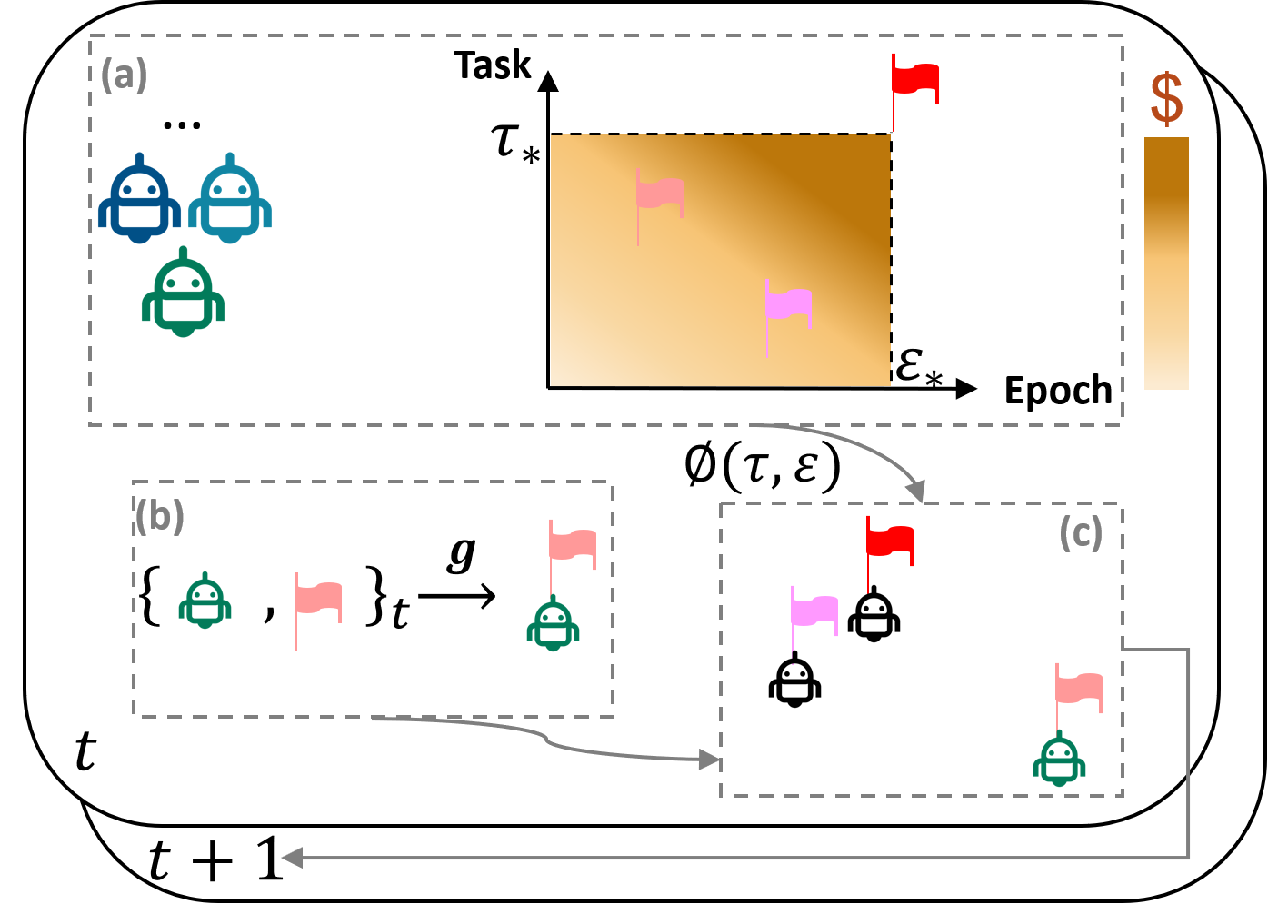}
\vspace{-2em}
\caption{Approach overview. (\textrm{a}) robot space and fidelity space; (\textrm{b}) evaluation on a selected design and a fidelity vector; (\textrm{c}) fitting a surrogate model with multi-fidelity evaluations.}
\label{fig: approch overview}
\end{figure}

The computational cost of evaluating a robot design depends on the target task and assigned epochs for learning a controller. 
\Sha{It takes more trials for a robot to accomplish or learn a high difficulty  task than one of low difficulty. For example, a robot would spend more iterations of trial and error to successfully navigate to a farther destination than a near one or to push an object to a more distant goal location than a near one.}

This motivates us to design a method that can utilize cheap approximate evaluations from easier tasks and fewer epochs of policy learning to guide search over the robot morphology space. The evaluations on non-target tasks or fewer epochs are referred to as \textit{low-fidelity evaluations} since they are approximations to the target evaluations. These low-fidelity evaluations are cheap yet relevant to the target evaluations. Therefore, low-fidelity evaluations can be leveraged to escape non-promising regions in morphology space without paying huge expense, and the budget then is reserved for identifying smaller promising regions.  Our approach towards sample-efficient and resource-efficient robot morphology optimization is shown in Fig.~\ref{fig: approch overview}. In the following subsections, we introduce a novel \textit{surrogate model} that can accurately fit a history of evaluations from different fidelity levels, and a selection criterion for deciding which fidelity evaluations to query.

\begin{figure}
\begin{algorithm} [H] 
\small
\captionsetup{font=small}
\caption{Neural fidelity warping for Multi-fidelity BO}
\textbf{Input:} $\mathcal{X}$, $\mathcal{Z}$, $\vect{z}_{\star}$, $c(\vect{z})$ and $budget$ \\
\textbf{Output:} $\vect{x^{\ast}}$
\begin{algorithmic}[1]

\State Initialize $\mathcal{D}_{0}$ with initial designs and evaluations
\State $cost = 0$
\Repeat
\State Learn the embedding weights $\vect{\theta}_{e}$ and the GP hyperparameters $\vect{\theta}_{k}$ on data $\mathcal {D}_{t-1}$ by minimize Equation~\ref{equa: hyper parameter learning}
\State Choose $(\vect{x}_t, \vect{z}_t$) by maximizing the acquisition function in Equation~\ref{equa: MF-ES}
\State Evaluate $(\vect{x}_{t}, \vect{z}_{t})$ to obtain $y_t= g(\vect{x}_{t}, \vect{z}_{t})$
\State $\mathcal {D}_{t} \leftarrow \mathcal {D}_{t-1} \bigcup \left \{ (\vect{x}_{t}, \vect{z}_{t}, y_t) \right \}$
\State $cost  \leftarrow cost + c(\vect{z}_t)$
\Until{$cost > budget$} 
\State \Return best $g(\vect{x}, \vect{z}=\vect{z}_{\star})$ recorded and the $\vect{x^{\ast}}$
\end{algorithmic}
\label{alg:nnwarping-mfbo}
\end{algorithm}
\vspace{-1em}
\end{figure}

\subsection{Neural Fidelity Warping Multi-fidelity Gaussian Process}

The key challenge of Multi-fidelity Bayesian Optimization (MFBO) is to build a \textit{surrogate model} that can fit data from different fidelity evaluations and predict outcomes of unseen robot morphologies at an arbitrary fidelity level. The \textit{surrogate model} of standard BO is for a single fidelity (i.e., \Sha{evaluations only by learning the target task with fixed epochs}). 

Since our goal is to search a robot morphology by its performance only on the target evaluation function, we consider the evaluations performed by a $[\tau, \varepsilon]$ where $\tau \neq \tau_{\star}$ and $\varepsilon \neq \varepsilon_{\star}$ as of low fidelity compared to evaluations performed by a $[\tau_{\star}, \varepsilon_{\star}]$. Low-fidelity evaluations are related to the target evaluation; hence, exploring the morphology space through low-fidelity evaluations can inform the target one's optimal location. Formally, we define a function $g: (\mathcal{X}\times \mathcal{Z}) \rightarrow \mathbb{R}$, where $\mathcal{Z} \subseteq \mathbb{R}^{2} $ represents the two dimensional space of task and number of epochs and $\vect{z} = [\tau, \varepsilon] \in \mathcal{Z}$ is termed as fidelity vector. The space of tasks can be specified as the time horizon for locomotion, the distance to a goal position for robot pushing, etc. 
The target $f$ which we aim to maximize is related to $g$ by $f(\vect{x}) = g(\vect{x},\vect{z}_{\star})$, where $\vect{z}_{\star}=[\tau_{\star}, \varepsilon_{\star}]$. 
Instead of maintaining a dataset of past evaluations as  $\mathcal{D}_{T}=\left \{ \left (\vect{x}_{t}, y_{t} \right ) \right \}_{t=0}^{T}$, here we store past evaluations of different fidelity vectors as $\mathcal{D}_{T}=\left \{ \left (\vect{x}_{t}, \vect{z}_{t}, y_{t} \right ) \right \}_{t=0}^{T}$ where $\vect{z}_{t}$ represents the fidelity vector queried for $t$-th evaluation.   

Adding the fidelity vector as an additional input to the \textit{surrogate model} for MFBO has been previously studied by~\cite{swersky2013multi, kandasamy2017multi, klein2017fast}. \cite{swersky2013multi} study MFBO in a limited setting where the fidelity space is a discrete finite set. In contrast, we follow the setting from~\cite{kandasamy2017multi, klein2017fast} that propose to utilize approximate evaluations by continuous fidelity space.
Restricting access to a finite set of low-fidelity evaluations prevents the benefits of exploring \sh{a broad range of} low-fidelity functions \sh{when a continuous spectrum of approximations is available.} 
\comment{to locate the optima efficiently.} Furthermore, it is nontrivial to pick a set of discrete low-fidelity functions since the fidelity vector is naturally defined in a continuous space for many problems. In a robot morphology design problem, for example, the target evaluation could be defined by a locomotion task of learning to move forward by \Sha{1000 steps with 1000 training epochs, i.e., $\vect{z}_{\star} =[1000, 1000]$. It is unclear if exploring $g(\cdot, z=[900, 1000])$ can help to unravel more information about the optima of the target $f=g(\cdot, \vect{z}_{\star})$ than exploring $g(\cdot, z=[800, 1000])$ or $g(\cdot, z=[1000, 800])$. Thus a $g$ over a continuous fidelity space $\mathcal{Z}=[0, 1000] \times [0, 1000]$ is more reasonable than a finite discrete set.}



As in standard BO, we assume $g$ is sampled from a GP model, i.e., $g\sim \mathcal{GP}(m, k)$, where $m: \mathcal{X}\times \mathcal{Z}\rightarrow \mathbb{R} $ and $k:(\mathcal{X}\times \mathcal{Z})^2\rightarrow \mathbb{R}$. Since the GP here models a multi-fidelity function, it is named Multi-fidelity Gaussian Process (MFGP). The kernel function $k$ defines the covariance between evaluations of morphology-fidelity pairs. Following~\cite{kandasamy2017multi, klein2017fast}, we use a factorized kernel $k([\vect{x}, \vect{z}], [\vect{x}{'}, \vect{z}{'}]) = k_x(\vect{x}, \vect{x}{'})k_z(\vect{z}, \vect{z}{'})$. $k_x$ measures the relationships between $\vect{x}$ and $k_{z}$ measures the relationships between $\vect{z}$. \cite{kandasamy2017multi} adopt a ARBF kernel (see Equation~\ref{equa: RBF-ARD}) for both $k_x$ and $k_z$, and~\cite{klein2017fast} adopt a Matérn kernel for $k_x$ and a finite-rank kernel for $k_z$.

Previous works~\cite{kandasamy2017multi, klein2017fast} fail to consider the complex relationships between different fidelity evaluations. This results from the limited kernels they adopt. Both ARBF kernel and Matérn kernel use an assumption of stationarity, which means they only depend on the distance between inputs and are invariant to the translations in inputs.  However, the assumption of stationarity is not true in the robot morphology design problem. The distance between fidelity level can not well capture the covariance between the corresponding evaluations. For example, the covariance between evaluations by 1000 training iterations and evaluations by 900 iterations should not be treated as equal as the covariance between 200 training iterations and 100 iterations. \Sha{The performance gap between final stages (e.g., 900 iterations and 1000 iterations) of learning can be less significant than the gap between initial stages (e.g., 100 iterations and 200 iterations).} Adopting the commonly used stationary kernels can mislead the search over the fidelity space.
Previous works~\cite{snoek2014input, assael2014heteroscedastic, calandra2016manifold, martinez2017bayesian} have tackled the non-stationary issue of a single fidelity setting, though don't consider the non-stationarity over the fidelity space. Moreover, since the non-stationary properties are not known a priori, simple operations such as transforming the fidelity using logarithm, which is usually used by searching learning rates for machine learning algorithms~\cite{bergstra2012random}, is not a general method. 

We address the non-stationarity by a neural network parameterized embedding function $\phi_e(\cdot; \vect{\theta}_{e})$ to embed the fidelity vector $\vect{z}$ into a representation $\vect{r}$. The goal of the embedding function is to embed the original fidelity to a space where the distance  can well represent the covariance between evaluations. The embedding weights $\vect{\theta}_{e}$ are jointly learned with the kernel parameters $\vect{\theta}_{k}$. Now we describe the learning and inference for the neural fidelity warping MFGP.

\mypara{Hyperparameters Learning $\vect{\theta}=[\vect{\theta}_{k}, \vect{\theta}_{e}]$}. $\vect{\theta}$ are learnt by minimizing the Negative Log Marginal Likelihood:
\begin{equation} 
\begin{aligned}
-\mathrm{log} \hspace{0.2em} p(\vect{Y}|\vect{X}, \vect{\theta}) = \frac{1}{2}\vect{Y}^{\top}(\vect{K} +  \sigma_{noise}^{2}\vect{I})^{-1}\vect{Y} +\\ \frac{1}{2} \mathrm{log}|\vect{K} + \sigma_{noise}^{2}\vect{I}|
\end{aligned}
\label{equa: hyper parameter learning}
\end{equation}
where \Sha{$\vect{Y} \in \mathbb{R}^{T}$ is a $T$ dimensional vector with $\vect{Y}[t]= y_{t}$, 
$\vect{X} \in \mathbb{R}^{T\times n}$ is a matrix of dimension $T \times n$ whose $t$-th row element equals $\vect{x}_{t}$,
and $\vect{K}\in \mathbb{R}^{T \times T}$ is a matrix of dimension $T\times T$ with $\vect{K}[i, j] = k([\vect{x}_{i}, \phi_e(\vect{z}_{i})], [\vect{x}_{j}, \phi_e(\vect{z}_{j})])$}

\mypara{Inference over unseen morphology-fidelity pair}. Denote $y$ as the predicted outcome of evaluating $(\vect{x}, \vect{z})$, then the posterior probability density of $y$ is also a Gaussian distribution: $y|\mathcal{D}_{T}, \vect{\theta} \sim \mathcal{N}(u, \sigma^{2})$:
\begin{equation} \label{equa: posterior distribution for xz}
\begin{gathered}
u(\vect{x}, \vect{z}) =  \vect{k}^{\top}(\vect{K} + \sigma_{noise}^{2}\vect{I})^{-1}\vect{Y}, \\
\sigma^{2}(\vect{x}, \vect{z}) =k([\vect{x}, \phi_e(\vect{z})], [\vect{x}, \phi_e(\vect{z})]) -\vect{k}^{\top}( \vect{K} +\sigma_{noise}^{2}\vect{I})^{-1}\vect{k},
\end{gathered}
\end{equation}
where $\vect{k}\in \mathbb{R}^{T}$ with $\vect{k}_{t} = k([\vect{x}, \phi_e(\vect{z})], [\vect{x}_{t}, \phi_e(\vect{z}_{t}]))$.




\subsection{Maximum Entropy reduction per cost for morphology and fidelity selection}
\Sha{In the multi-fidelity setting, each evaluation's observation is determined by a triplet of the task, the number of learning epochs, and a candidate robot morphology. As in standard BO, selection criteria should be set to compute the next robot morphology and next task and number of epochs.}
We follow~\cite{klein2017fast} which adopt a Multi-fidelity entropy search acquisition function. The next-to-evaluate $(\vect{x}, \vect{z})$ is aimed to acquire more information about the location of $\vect{x}_{*}^{(z_{\star})}$, the global optima of $f$. Given $\mathcal{D}_{T}$, the information about $\vect{x}_{*}^{(z_{\star})}$ can be measured in terms of the negative differential entropy of $\vect{x}_{*}^{(z_{\star})}$. 
Therefore, the selection strategy is to determine next $(\vect{x}, \vect{z})$ which can maximize the expected reduction in this conditional differential entropy per cost.
\begin{equation}
\begin{aligned}
a(\vect{x}, \vect{z}) = & \frac{1}{c(\vect{z})}\{  H(\vect{x}_{*}^{(z_{\star})}|\mathcal{D}_{T}) - \\& \E_{P( y_x^{(z)}|\mathcal{D}_{T}, (\vect{x}, \vect{z}) )} H(\vect{x}_{*}^{(z_{\star})}|\mathcal{D}_{T} \cup  \{ (\vect{x}, \vect{z}, y_x^{(z)})  \} ) \},\\
\end{aligned}
\label{equa: MF-ES}
\end{equation}
where $\vect{x}_{*}^{(z_{\star})}:= \argmax_{x\in \mathcal{X}}g(\vect{x}, {\vect{z}_{\star})}:=  \argmax_{x\in \mathcal{X}}f(\vect{x})$, and $y_x^{(z)}$ is the observation obtained by evaluating $\vect{x}$ at fidelity $\vect{z}$. $H(\vect{x}_{*}^{(z_{\star})}|\mathcal{D}_{T})$ is the differential entropy of $\vect{x}_{*}^{(z_{\star})}$ conditioned on $\mathcal{D}_{T}$, and the expectation is taken over the posterior predictive distribution of $y_x^{(z)}$. 

\Sha{The full algorithm is summarised in Algorithm~\ref{alg:nnwarping-mfbo}. In each iteration, the hyperparameters are first learned with past evaluation data, then the next robot morphology and fidelity vector are computed by maximizing the acquisition function. The algorithm terminates when the cost budget is exhausted.}

\section{Experimental Results}

\Sha{We perform experiments on two robot morphology design tasks to verify the efficacy of our method. In this section we first provide implementation details and then compare our method to standard baselines.} 
\label{sec:result}

\begin{figure*}

  \begin{subfigure}{0.49\linewidth}
  \includegraphics[width=0.198\linewidth]{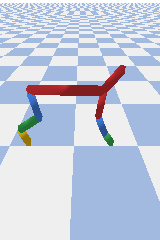}\hfill
  \includegraphics[width=0.198\linewidth]{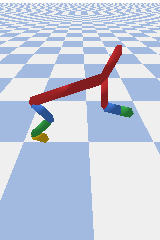}\hfill
  \includegraphics[width=0.198\linewidth]{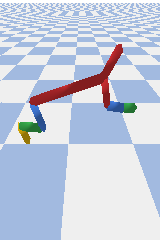}\hfill
  \includegraphics[width=0.198\linewidth]{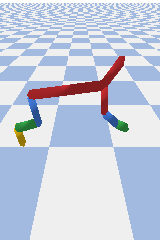}\hfill 
  \includegraphics[width=0.198\linewidth]{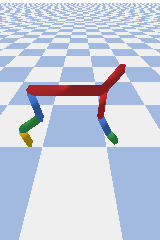}
  \caption{a Cheetah designed by a human}
  \label{fig: cheetah human}
  \end{subfigure} 
  \hspace{0.2em}
  \begin{subfigure}{0.49\linewidth}
  \includegraphics[width=0.198\linewidth]{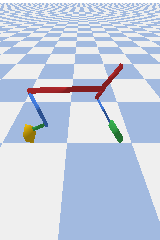}\hfill
  \includegraphics[width=0.198\linewidth]{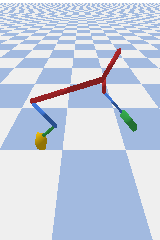}\hfill
  \includegraphics[width=0.198\linewidth]{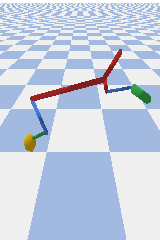}\hfill
  \includegraphics[width=0.198\linewidth]{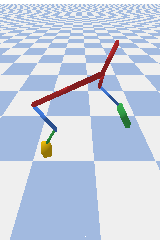}\hfill 
  \includegraphics[width=0.198\linewidth]{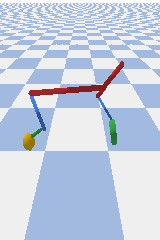}
  \caption{a Cheetah optimized by our method}
  \label{fig: cheetah opt}
  \end{subfigure}\par\medskip
  \vspace{-0.5em}
  \begin{subfigure}{0.49\linewidth}
  \includegraphics[width=0.198\linewidth]{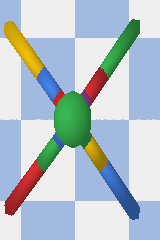}\hfill
  \includegraphics[width=0.198\linewidth]{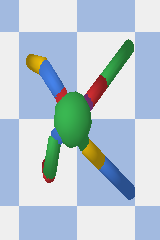}\hfill
  \includegraphics[width=0.198\linewidth]{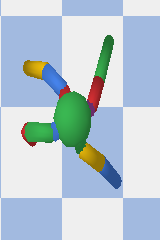}\hfill
  \includegraphics[width=0.198\linewidth]{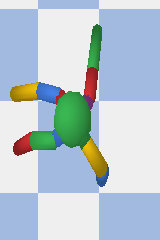}\hfill
  \includegraphics[width=0.198\linewidth]{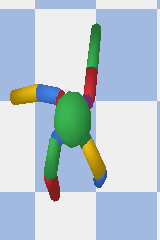}
  \caption{an Ant designed by a human}
  \end{subfigure} 
  \hspace{0.2em}
  \begin{subfigure}{0.49\linewidth}
  \includegraphics[width=.198\linewidth]{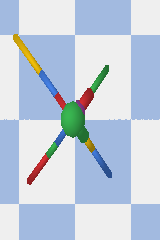}\hfill
  \includegraphics[width=.198\linewidth]{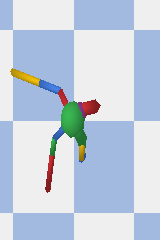}\hfill
  \includegraphics[width=.198\linewidth]{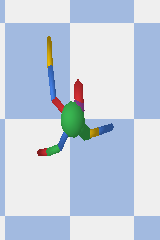}\hfill
  \includegraphics[width=.198\linewidth]{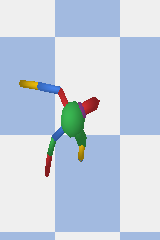}\hfill 
  \includegraphics[width=.198\linewidth]{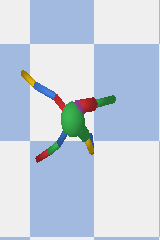}
   \caption{an Ant optimized by our method}
   \label{fig: ant opt}
  \end{subfigure}\par\medskip
  \vspace{-1em}
  \caption{$(\textrm{a}), (\textrm{b})$: Side view: Cheetah performing cantering. $(\textrm{c}), (\textrm{d})$: Top-down view: Ant performing crawling.}
  \vspace{-1.5em}
\end{figure*}

\subsection{Implementation Details and Baselines}
The embedding function for the fidelity vector is $\phi_e(\cdot)$, which is a multi-layer perceptron with Sigmoid activations, and the hidden units of $\phi_e(\cdot)$ are of dimensions (2, 6, 2). We implemented NFW-MFBO in GPy\footnote{https://github.com/SheffieldML/GPy}, an open-sourced GP framework. We adopt L-BFGS \cite{liu1989limited} for learning and DIRECT~\cite{jones1993lipschitzian} to optimize the acquisition function.  Optimization starts with initial designs computed by Latin hypercube sampling (LHS). For all the legged robots, we use PPO~\cite{schulman2017proximal} as the locomotion controller. We use 10 parallel threads and 256 mini batches per PPO epoch, and PPO parameters are trained using Adam~\cite{kingma2014adam} with a learning rate of $3\times10^{-4}$. The code and videos are available at \url{https://github.com/husha1993/FWBO}.

We compare our approach against two non-BO methods: (1) Random search; (2) Covariance Matrix Adaptation Evolution Strategy (CMA-ES); two single fidelity BO methods: (3) BO with entropy search  acquisition~\cite{hennig2012entropy}; (4) HPC-BBO~\cite{liao2019data}; and two MFBO methods: (5) BOCA~\cite{kandasamy2017multi} that adopts an ARBF kernel for $k_{z}$; (6) FABOLAS~\cite{klein2017fast} that adopts a finite-rank kernel for $k_{z}$. The main distinction between our approach and the two MFBO baselines lies in the \textit{surrogate model}, therefore we keep the acquisition function identical for a fair comparison. We also report the performance of the default morphologies from PyBullet denoted as Human.  

\comment{
\subsection{Optimization benchmark functions}
We start with two popular global optimization benchmark functions
to demonstrate that modeling non-stationarity in the fidelity space is critical to yield an effective multi-fidelity BO. (1) $\mathit{Branin}$ function~\cite{forrester2008engineering}; (2) $\mathit{Park}$ function~\cite{park1991tuning}: The input is four dimensional and each dimension is in $[0, 1]$. $f(\vect{x}) = $}



\subsection{Robot morphology design}
We consider two robot morphology design problems: Cheetah and Ant. Our environments are simulated with PyBullet~\cite{coumans2016pybullet}. We first describe the morphology space, action space, state space, and reward function for each environment. 

\noindent \textbf{Cheetah Env}: We consider a 13D morphology space for Cheetah where 6 dimensions indicate scaling factor of length of 6 limbs and 7 dimensions indicate scaling factor of radius of 6 limbs plus 1 torso. The morphology is hence represented by an element-wise multiplication between the design vector $\vect{x}$ and a unit design: $\vect{x} \circ [0.12,0.12,0.12,0.12,0.12,0.12,0.046,0.046, 0.046,0.046, \\ 0.046,0.046,0.046]$. The range of the design vector is the product space using the interval $x_{i}\in [0.3, 2]$ for each element.
The action space is 6D, representing 6D continuous torques at each of the hip, knee, and ankle on both front and back legs.  
The state space is 18D and consists of pose, and speed of joints and torso.
The goal of the Cheetah is to move along the $x$-direction as fast as possible, and the reward is the speed along the $x$-direction. Every episode terminates when the robot torso touches the ground or after a maximum number of timesteps. We report the performance by the return averaged over steps. We set the target evaluation as one of 250 training epochs, and 250 steps. The fidelity vector $\vect{z}=[\tau, \varepsilon]$ uses the interval $\tau, \varepsilon \in [0, 1]$. The low-fidelity evaluations are of training epochs computed by $50+ \varepsilon \times 200$, and task length by $50+ \tau \times 200$. We consider the FLoating point OPerations (FLOPs) as the cost per evaluation since FLOPs do not depend on the implementation platforms. The FLOPs for each evaluation is proportional to the number of learning epochs, thus we define the cost function as $c(\vect{z}) = \frac{1}{250} (50+ \varepsilon \times200)$. The budget is set to 50. 

\noindent \textbf{Ant Env}: We consider a 25D morphology space for Ant where 12 dimensions indicate the absolute length of 12 limbs and 13 dimensions indicate absolute radius of 12 limbs plus 1 head. The action space is 12D, representing 12D torques at each joint. The state space is 28D and consists of pose, and speed of joints and head. The reward is the distance travelled along the $x$-direction averaged over steps. The \comment{termination condition,}definition of fidelity space, the target evaluation function, the cost function and the budget are the same as Cheetah. 

\mypara{Quantitative Evaluation.} Table~\ref{tab: results} reports the best morphology's performance on the target evaluation returned by baselines and our method. For Cheetah, the initial designs are obtained by sampling 6 designs through LHS for single fidelity baselines. For MFBO methods, the initial designs are obtained by sampling 10 designs through LHS, sampling 10 fidelity vectors, and then randomly concatenating them to produce 10 pairs. The cost for initial designs for single fidelity baselines and MFBO is 6, 6.55, respectively. For Ant, the number of initial designs is 15 for single fidelity baselines and is 25 for MFBO, and the costs are 15, 15.99, respectively. The numbers of initial designs are chosen so that the initial budget is roughly the same for all methods.

Our method can find a morphology whose performance exceeds the morphologies returned by all baselines. As expected, CMA-ES and standard BO, which only evaluate candidate morphologies by an expensive target evaluation function, can waste some budget in exploring non-promising morphology regions and end with a sub-optimal return. HPC-BBO performs worse than standard BO. This may result from HPC-BBO's reliance on hallucinated evaluations to predict a batch of candidate morphologies. Incorrect hallucinations can be a risk in wasting budget on a batch of evaluations. Such risk would be more serious when the design space is high dimensional, where hallucinating evaluations can be even more challenging. In contrast, our method only predicts one morphology each step, hence reduces the impact of an imperfect surrogate model.

By comparing our method to the other two MFBO baselines, we can validate the importance of modeling non-stationarity across the fidelity space.  FABOLAS~\cite{klein2017fast} intends to incorporate a strong prior that a machine learning algorithm's performance usually increases with more training data by adopting a finite-rank kernel of linear basis functions. However, such a prior is not flexible enough to capture complex non-stationarity compared to our neural fidelity warping method.  FABOLAS~\cite{klein2017fast} performs worse in the two design problems than our method and even worse  in Ant than BOCA~\cite{kandasamy2017multi}, which ignores the non-stationarity.

\begin{table}[htb]
\centering
\begin{tabular}{@{}l|c|c@{}}
\toprule 
Method       & Cheetah       & Ant \\
\hline
Human  &  $0.352$   & $0.460$\\
\hline
Random      &  $0.367 \pm 0.017$  & $ 1.280 \pm 0.130$ \\
CMA-ES     &  $0.368\pm 0.051$  & $1.205 \pm 0.211$ \\ 
\hline
BO~\cite{hennig2012entropy} &  $0.462 \pm 0.007$  &$ 1.390 \pm 0.091$ \\
HPC-BBO~\cite{liao2019data} &  $0.413 \pm 0.018$   & $ 1.337 \pm 0.131$  \\ 
\hline
BOCA~\cite{kandasamy2017multi} &  $0.526 \pm 0.123$   & $1.511 \pm 0.093$ \\
FABOLAS~\cite{klein2017fast} & $0.552 \pm 0.133$ & $ 1.398 \pm 0.169$ \\ 
NFW (Ours)  & $\mathbf{0.717} \pm  0.101$  & $\mathbf{1.871} \pm 0.117$ \\

\bottomrule
\end{tabular}
\vspace{-3pt}
\caption{Quantitative results in two robots morphology design tasks. $\pm$ indicates standard deviation measured using three independently seeded optimization  runs.}
\label{tab: results}
\vspace{-0.5em}
\end{table}

\mypara{Qualitative Evaluation.} We further investigate the optimized morphologies. For \textbf{Cheetah Env}, the optimal design found by our method 
has a muscular back thigh that can support Cheetah to lift its front limbs to a very high position (Fig.~\ref{fig: cheetah opt}) to obtain massive momentum to perform a leap. In contrast, the default Cheetah from PyBullet has equal front thigh and back thigh, and it cannot raise its front limbs to as high a position as our optimal Cheetah. \comment{Therefore, our Cheetah can run at a faster speed.} For \textbf{Ant Env}, the optimal design found by our method has a light-weighted head and a long thin back leg functioning like a tail (Fig.~\ref{fig: ant opt}). In contrast, the default Ant from PyBullet has four legs of the same length and radius. By sweeping the long back leg, the optimal Ant can swiftly crawl forward. 


\section{Conclusion}
We contribute to methods for automatic robot design by developing a continuous multi-fidelity Bayesian Optimization framework that efficiently utilizes computational resources via low-fidelity evaluations.  By considering the non-stationarity across fidelities, our method can leverage information from multi-fidelity evaluations to accurately estimate the target-fidelity objective, leading to better robot design optimization results within a limited computational resource budget.

\bibliographystyle{IEEEtran}
\bibliography{icra}

\end{document}